\documentclass[letterpaper, 10 pt, conference]{ieeeconf}  % Comment this line out if you need a4paper
\usepackage[T1]{fontenc}
\usepackage{amssymb}

\usepackage{amsmath}
\usepackage{array}
\usepackage{cite}

\usepackage{makecell}
\usepackage{booktabs}
\usepackage{wrapfig}
\usepackage{tcolorbox}
\usepackage{algorithm}
\usepackage[noend]{algpseudocode}
\usepackage{tabularray} % For the tblr environment
\usepackage{tcolorbox}  % For colored boxes and styling
\usepackage{xcolor}     % For text coloring
\usepackage{graphicx} % for including images

\usepackage{caption}

\usepackage{multirow} 

\usepackage{xspace}
\usepackage{hyperref}
\usepackage{cleveref}
\usepackage{varwidth}
\usepackage{arydshln}
\usepackage{xcolor}

\newcommand{\ALGname}{QoQ\xspace}
\newcommand{\Algname}{\textbf{Q}uality \textbf{o}ver \textbf{Q}uantity\xspace}
\newcommand{\algname}{Quality over Quantity\xspace}
\newcommand{\stdv}[1]{\scalebox{.80}{~$\pm$~#1}}

\IEEEoverridecommandlockouts                              % This command is only needed if 
\title{\LARGE \bf
Quality over Quantity: Demonstration Curation via Influence Functions for Data-Centric Robot Learning
}
\author{Haeone Lee$^{* 1}$, Taywon Min$^{1}$ Junsu Kim$^{1}$, Sinjae Kang$^{1}$, Fangchen Liu$^{2}$, Lerrel Pinto$^{3}$, and Kimin Lee$^{1}$
\thanks{$^{1}$KAIST $^{2}$University of California, Berkeley $^{3}$New York University.}
\thanks{$*$ Correspondence to \texttt{haeone.lee@kaist.ac.kr}}
}

\begin{document}

\maketitle
\thispagestyle{empty}
\pagestyle{empty}

%%%%%%%%%%%%%%%%%%%%%%%%%%%%%%%%%%%%%%%%%%%%%%%%%%%%%%%%%%%%%%%%%%%%%%%%%%%%%%%%
\begin{abstract}
    Learning from demonstrations has emerged as a promising paradigm for end-to-end robot control, particularly when scaled to diverse and large datasets. 
    However, the quality of demonstration data, often collected through human teleoperation, remains a critical bottleneck for effective data-driven robot learning. 
    Human errors, operational constraints, and teleoperator variability introduce noise and suboptimal behaviors, making data curation essential yet largely manual and heuristic-driven.
    In this work, we propose \Algname (\textbf{\ALGname}), a grounded and systematic approach to identifying high-quality data by defining data quality as the contribution of each training sample to reducing loss on validation demonstrations. 
    To efficiently estimate this contribution, we leverage influence functions, which quantify the impact of individual training samples on model performance.
    We further introduce two key techniques to adapt influence functions for robot demonstrations: (i) using maximum influence across validation samples to capture the most relevant state-action pairs, and (ii) aggregating influence scores of state-action pairs within the same trajectory to reduce noise and improve data coverage.
    Experiments in both simulated and real-world settings show that \ALGname consistently improves policy performances over prior data selection methods.
    % It includes its ability to curate high-quality data from in-the-wild datasets such as DROID, highlighting \ALGname's effectiveness across diverse real-world environments.
    % We further demonstrate its ability to curate high-quality data from in-the-wild datasets such as DROID, highlighting its effectiveness across diverse real-world environments. 
\end{abstract}

%%%%%%%%%%%%%%%%%%%%%%%%%%%%%%%%%%%%%%%%%%%%%%%%%%%%%%%%%%%%%%%%%%%%%%%%%%%%%%%%
\section{INTRODUCTION}
Learning from demonstrations has shown potential in end-to-end robot control, particularly when scaling both the diversity and quantity of demonstration data~\cite{walke2023bridgedata,o2024open, khazatsky2024droid, black2410pi0, bjorck2025gr00t, kim24openvla, team2024octo}. 
However, the quality of robot demonstration data, commonly collected through human teleoperation, significantly impacts performance when training with supervised learning methods like behavior cloning (BC)~\cite{belkhale2023data, mandlekar2022matters}.
Human errors, operational constraints, and varying skill levels of teleoperators introduce noise and suboptimal behaviors into these datasets, making effective curation critical for successful data-driven robot learning.

Despite its importance, data curation remains largely manual, expensive, and reliant on heuristic judgments. 
% Despite its importance, data curation remains largely manual, expensive, and reliant on heuristic judgments. The definition of ``high-quality data''  itself varies widely across applications and research objectives, raising a fundamental question:
% \begin{center}
% \textit{What constitutes high-quality data, and how can we systematically identify such data?}
% \end{center}
Previous approaches have attempted to address this challenge using proxy metrics, including similarity to expert demonstrations~\cite{du2023behavior, linflowretrieval} and mutual information between state and action distributions~\cite{hejna2025robot}. 
However, these metrics often fail to capture which training data truly contributes to improved policy performance.
%by relying on manually engineered metrics.
% However, these metrics often fail to capture which training data truly contributes to improved policy performance.

\begin{figure*}[t]
    \centering
    \includegraphics[width=0.9\textwidth]{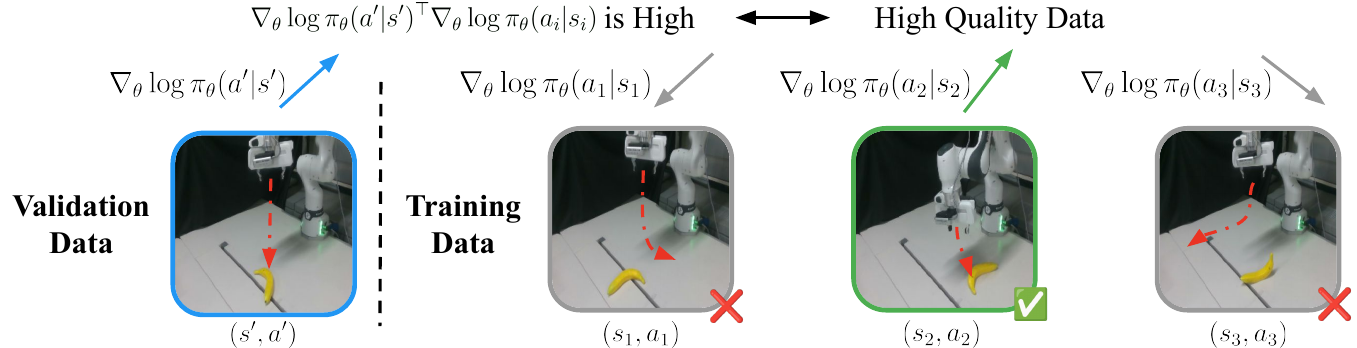}
    \caption{\textbf{Illustration of high-quality data.}
    Our robot data curation method, \ALGname, selects trajectories based on \emph{their direct contribution to policy performance}, using influence functions~\cite{koh2017understanding} to quantify this impact. 
    Specifically, we measure the similarity between gradients of validation data (blue) and those of training data; higher similarity (green) indicates that including a particular state-action pair will effectively reduce validation loss.
    By prioritizing these high-impact data points, we systematically identify and preserve the most valuable training data that drive performance improvement.}
    \label{fig:concept}
    \vspace{-5pt}
\end{figure*}

In this work, we call for a data curation framework based on influence functions~\cite{koh2017understanding}. Influence functions can measure the contribution of training data to reducing loss on a small set of validation demonstrations that represent the target desired behavior (see Figure~\ref{fig:concept}).
% \textcolor{red}{
% In this work, we propose a more grounded definition of data quality: the contribution of training data to reducing loss on a small set of carefully curated validation demonstrations.
%To efficiently estimate this contribution, we leverage influence functions~\cite{koh2017understanding}, which estimate how individual training samples affect model performance (see Figure~\ref{fig:concept}).
%\textcolor{blue}{affect a model's predictive accuracy on validation state-action pairs.} 
We remark that performance on unseen validation samples serves as a useful metric for a policy’s generalization capabilities, thereby allowing our definition to capture the complex relationship between training data and policy performance.

% We remark that performance on unseen validation samples serves as a proxy for a policy’s generalization capabilities, thereby allowing our definition to capture the complex relationship between training data and policy performance.

%%%% paragraph 4
However, we find that naively applying influence functions to robot demonstrations yields noisy signals and tends to select redundant state-action pairs, resulting in poor coverage of the state space (see Section~\ref{subsec:ablation}). 
To address these issues, we propose \Algname (\textbf{\ALGname}), which introduces two key techniques for effectively applying influence functions to robot demonstrations:
First, we measure each state-action pair's influence by its maximum influence across validation samples, focusing only on the most relevant validation state-action pair rather than averaging across all validation data. 
Second, we implement trajectory-wise curation, which aggregates influence scores of state-action pairs within the same trajectory and then selects high-quality trajectories based on the aggregated scores. 
This approach significantly reduces noise in the influence signal while ensuring broad state coverage, capturing diverse and informative robot behaviors.

%%%% paragraph 5
We evaluate \ALGname on both Robomimic simulation~\cite{mandlekar2022matters} and multiple real-robot manipulation tasks. 
Our experiments show that \ALGname successfully filters out low-quality demonstrations (\textit{e.g.,} failure cases), significantly improving policy performance when trained on curated datasets.
By considering the true effect to the policy in data curation, \ALGname substantially outperforms the baseline methods that rely on state and action features~\cite{du2023behavior, linflowretrieval} up to 23.2\% in Robomimic simulation and 30.0\% in real robot success rate. 
%\textcolor{red}{We also show that \ALGname is able to leverage policy rollout data as a validation set to curate a train dataset that can achieve high performance afterwards.}
We further demonstrate that \ALGname can curate high-quality data from the DROID~\cite{khazatsky2024droid} dataset, which was collected in the wild and encompasses diverse environments and object locations.

% Our ablation studies also verify that each component of our method is critical for effective data curation.

%%%%%%%%%%%%%%%%%%%%%%%%%%%%%%%%%%%%%%%%%%%%%%%%%%%%%%%%%%%%%%%%%%%%%%%%%%%%%%%%
%%%%%%%%%%%%%%%%%%%%%%%%%%%%%%%%%%%%%%%%%%%%%%%%%%%%%%%%%%%%%%%%%%%%%%%%%%%%%%%%
\section{Related Work}

\subsection{Data valuation}

Data valuation aims at quantifying the contribution of individual data points to the performance of machine learning models.
One approach is Data Shapely~\cite{ghorbani2019data, jia2019efficient, kwon2021beta}, which leverages cooperative game theory to data valuation by computing the average marginal contribution of each datapoint to the models' performance across all possible subsets of training data. 
Influence functions~\cite{koh2017understanding} offer another approach by estimating the effect of upweighting or removing a training point on the models' parameters and loss, without requiring full retraining.
This is achieved via first-order approximations involving gradients and inverse Hessian-vector products, enabling efficient estimation. 
% While influence functions have been widely explored in vision or language tasks~\cite{pruthi2020estimating, kwon2024datainf, lin2024token, grosse2023studying, min2025understanding}, our work is the first to apply influence functions in the context of robot learning.

\subsection{Robot data curation}
To filter low-quality trajectories, previous work proposed their own definition of the quality of robot data. 
For example, retrieval-based approaches~\cite{du2023behavior, nasiriany2022learning, memmel2025strap} define it as similar points with expert data in feature space.
Flow retrieval~\cite{linflowretrieval} extends this idea by representing trajectories using optical flow to better capture temporal dynamics. 
DemInf~\cite{hejna2025robot} curates the training dataset by selecting trajectories with high mutual information between states and actions as a proxy for demonstration quality. 
Demo-SCORE~\cite{chen2025curating} curates trajectories that are reproducible for the robot to imitate; it filters them using a classifier trained on policy rollouts to identify trajectories that consistently lead to task success.
Compared to prior work, we define robot data quality based on the \emph{direct performance contribution} to the learned policy. 

% Related to our work, DataMIL~\cite{dass2025datamil} and CUPID~\cite{agia2025cupid} also attribute robotic data contributions using influence functions. 
% Unlike DataMIL and CUPID which average influence scores over entire validation transitions, we measure each state-action pair's influence by its maximum influence across validation samples, focusing only on the most relevant validation state-action pair.
% CUPID focuses on curating robot data using returns estimated from policy rollouts, while our method is generally more concerned about effectively utilizing the validation set for robotic data curation.

Related to our work, DataMIL~\cite{dass2025datamil} and CUPID~\cite{agia2025cupid} also attribute robotic data contributions using influence functions. 
%\textcolor{red}{
CUPID uses returns estimated from policy rollouts to curate robot data. When using only successful trajectories of return 1.0 as a validation set, CUPID score is equivalent to measuring each state-action pair's influence by its sum over the entire validation transitions.
Similarly, DataMIL uses the validation loss computed over the entire validation set.
However, we found that this leads to unstable influence score estimations as shown in our experiment. This is because not all validation transitions are helpful to evaluate the quality of state-action pairs of interest, since they could correspond to different behaviors (e.g., pick-and-place behavior might not help identify useful behavior for screwing).
% This is because not every state-action pair in the validation trajectory is relevant to train state-action pairs of interest in their behavior, which could introduce noise during influence estimation.
Therefore, we measure each state-action pair's influence by its maximum influence across validation samples, focusing only on the most relevant validation state-action pair.
%}
% not all validation transitions are helpful to evaluate the influence of the current state-action pair, as their states might be different. 
% mention CUPID's big approximations 
% mention DataMIL's average objectives
%  measure each state-action pair's influence by its maximum influence across validation samples, focusing only on the most relevant validation state-action pair rather than averaging across all validation data.
% CUPID focuses on curating robot data using returns estimated from policy rollouts, while our method is generally more concerned about effectively utilizing the validation set for robotic data curation.
% additionally comment why this is important (one of reviewer's commments)
% We refer readers to Appendix~\ref{sec:extended_related_work} for further comparison with prior works and techniques of leveraging suboptimal data in robot learning.

%%%%%%%%%%%%%%%%%%%%%%%%%%%%%%%%%%%%%%%%%%%%%%%%%%%%%%%%%%%%%%%%%%%%%%%%%%%%%%%%
\section{Preliminaries}

\subsection{Learning from Demonstrations}
We model the robot learning problem as a sequential decision-making process, where a policy outputs an action $a$ given a state $s$.
We consider a setting in which a robot learns from a dataset of demonstrations using Behavior Cloning (BC)~\cite{pomerleau1988alvinn}. 
The dataset $\mathcal{D}$ consists of trajectories, each representing a sequence of states and actions:
    \[
    \mathcal{D} = \{ \tau_i \}_{i=1}^{N}, \quad \text{where} \quad \tau_i = (s_0, a_0, s_1, a_1, \dots).
    \]

Given this dataset $\mathcal{D}$, we train the policy $\pi_\theta$ to imitate the demonstrated behavior by minimizing the BC loss function, defined as the negative log-likelihood of the demonstrated actions:
    \[
    \mathcal{L}_{\tt BC}(\mathcal{D}; \theta)=\mathbb{E}_{(s,a) \sim \mathcal{D}} \left[ -\log \pi_\theta(a | s) \right],
    \]
where $\theta$ denotes the parameters of the policy.

\subsection{Influence Functions}
\label{sec:pre_if}
% \begin{itemize}

% \item Use either SI (MKS) or CGS as primary units. (SI units are encouraged.) English units may be used as secondary units (in parentheses). An exception would be the use of English units as identifiers in trade, such as Ò3.5-inch disk driveÓ.
% \item Avoid combining SI and CGS units, such as current in amperes and magnetic field in oersteds. This often leads to confusion because equations do not balance dimensionally. If you must use mixed units, clearly state the units for each quantity that you use in an equation.
% \item Do not mix complete spellings and abbreviations of units: ÒWb/m2Ó or Òwebers per square meterÓ, not Òwebers/m2Ó.  Spell out units when they appear in text: Ò. . . a few henriesÓ, not Ò. . . a few HÓ.
% \item Use a zero before decimal points: Ò0.25Ó, not Ò.25Ó. Use Òcm3Ó, not ÒccÓ. (bullet list)

% \end{itemize}
Influence functions~\cite{koh2017understanding} approximates how the model parameters, or a function of the model parameters (\textit{e.g.,} validation loss) changes regarding a specific training point $(x_i, y_i)\in\mathcal{D}_{\tt tr}$. 

Specifically, given a loss function $\mathcal{L}$ and a training dataset $\mathcal{D}_{\tt tr}$, the $\varepsilon$-weighted risk minimizer for a single training data point $(x_i, y_i)$ is defined as $\theta^{(i)}(\varepsilon) := \underset{\theta \in \Theta}{\mathrm{arg\,min}} \ \frac{1}{|\mathcal{D}_{\tt tr}|} \sum_{(x,y) \in \mathcal{D}_{\tt tr}}\mathcal{L}(f_\theta(x), y) + \varepsilon \mathcal{L}(f_\theta(x_i), y_i)$.

This indicates that a single training sample, $(x_i, y_i)$, has been up-weighted by a small perturbation $\varepsilon$.
The influence function $\mathcal{I}_{\theta}(x_i, y_i)$ is defined as the derivative of $\theta^{(i)}(\varepsilon)$ at $\varepsilon=0$, 
quantifying how the model parameters change due to the data point $(x_i, y_i)$:

    \begin{equation}
    \mathcal{I}_\theta(x_i, y_i) := \left. \frac{d\theta^{(i)}(\varepsilon)}{d\varepsilon} \right|_{\varepsilon=0}.
    % \eqno{(1)} 
    \end{equation}
    
While this describes how model parameters shift, our primary goal is to quantify how each training sample influences validation loss. This can be achieved by applying the chain rule:

    \begin{equation}
    \mathcal{I}_{\tt val}(x_i, y_i) := \left. \nabla_\theta \mathcal{L}(\mathcal{D}_{\text{val}}; \theta)^\top \frac{d\theta^{(i)}(\varepsilon)}{d\varepsilon} \right|_{\varepsilon=0},
    % \eqno{(2)}
    \end{equation}
    
where $\mathcal{D}_{\tt val}$ denotes the validation dataset, and $\mathcal{L}(\mathcal{D}_{\tt val}; \theta)$ is the validation loss.
The influence function $\mathcal{I}_{\tt val}(x_i, y_i)$ estimate the degree of change in validation loss when a training sample $(x_i, y_i)$ is up-weighted. Under standard assumptions (twice-differentiability and strong convexity), this can be approximated as:
\begin{equation}
\label{eq:influence}
\scalebox{0.9}{$
\mathcal{I}_{\tt val}(x_i, y_i) :=
- \nabla_\theta \mathcal{L}(\mathcal{D}_{\tt val};\theta)^{\top}
H(\mathcal{D}_{\tt tr}; \theta)^{-1}
\nabla_{\theta}\mathcal{L}(x_i,y_i;\theta)
$}
\end{equation}
% \begin{equation}
%     \label{eq:influence}
%     \mathcal{I}_{\tt val}(x_i, y_i) :=  - \nabla_\theta \mathcal{L}(\mathcal{D}_{\tt val};\theta)^{\top} H(\mathcal{D}_{\tt tr}; \theta)^{-1} \nabla_{\theta}\mathcal{L}(x_i,y_i;\theta).
%     % \eqno{(3)}
% \end{equation}

$H(\mathcal{D}_{\tt tr}; \theta) := \nabla^2_\theta \mathcal{L}(\mathcal{D}_{\tt tr}; \theta)$ is the Hessian matrix of the training loss, and $\theta$ is the minimizer of the empirical risk over $\mathcal{D}_{\tt tr}$.
A lower value of $\mathcal{I}_{\tt val}(x_i, y_i)$ indicates the sample decreases the validation loss (potentially beneficial) and vice versa. 

However, computing the inverse Hessian in~\cref{eq:influence} is often computationally prohibitive. To address this, \cite{pruthi2020estimating} proposed a first-order approximation that omits the Hessian. Furthermore, they found that normalizing gradients improves the stability of influence estimation. With this modification, influence functions become:

    \begin{equation}
    \label{eq:influence_bc_norm}
     \mathcal{I}_{\tt val}(x_i, y_i) :=  - \nabla'_\theta \mathcal{L}(\mathcal{D}_{\tt val};\theta)^{\top} \nabla'_{\theta}\mathcal{L}(x_i,y_i;\theta),
     % \eqno{(4)}
    \end{equation}
    
    % \begin{align}
    % \mathcal{I}_{\tt val}(x_i, y_i) :=  
    % &- \nabla_\theta \mathcal{L}(\mathcal{D}_{\tt val};\theta)^{\top} \nonumber \\
    % &\cdot H(\mathcal{D}_{\tt tr}; \theta)^{-1} 
    % \nabla_{\theta}\mathcal{L}(x_i,y_i;\theta).
    % \end{align}
        
where the normalized gradient is defined as $\nabla'_\theta \mathcal{L} := \frac{\nabla_\theta \mathcal{L}}{\left\| \nabla_\theta \mathcal{L} \right\|_2}$. In our work, we utilize \cref{eq:influence_bc_norm} to estimate influence values on robot datasets.

%%%%%%%%%%%%%%%%%%%%%%%%%%%%%%%%%%%%%%%%%%%%%%%%%%%%%%%%%%%%%%%%%%%%%%%%%%%%%%%%
% \subsection{Equations}

% The equations are an exception to the prescribed specifications of this template. You will need to determine whether or not your equation should be typed using either the Times New Roman or the Symbol font (please no other font). To create multileveled equations, it may be necessary to treat the equation as a graphic and insert it into the text after your paper is styled. Number equations consecutively. Equation numbers, within parentheses, are to position flush right, as in (1), using a right tab stop. To make your equations more compact, you may use the solidus ( / ), the exp function, or appropriate exponents. Italicize Roman symbols for quantities and variables, but not Greek symbols. Use a long dash rather than a hyphen for a minus sign. Punctuate equations with commas or periods when they are part of a sentence, as in

% $$
% \alpha + \beta = \chi \eqno{(1)}
% $$

% Note that the equation is centered using a center tab stop. Be sure that the symbols in your equation have been defined before or immediately following the equation. Use Ò(1)Ó, not ÒEq. (1)Ó or Òequation (1)Ó, except at the beginning of a sentence: ÒEquation (1) is . . .Ó

%%%%%%%%%%%%%%%%%%%%%%%%%%%%%%%%%%%%%%%%%%%%%%%%%%%%%%%%%%%%%%%%%%%%%%%%%%%%%%%%
\section{\algname (\ALGname)}

In this section, we introduce \Algname (\textbf{\ALGname}), a method for curating high-quality robot data using influence functions.
In Section~\ref{subsec:criteria}, we define what constitutes high-quality data. 
In Section~\ref{subsec:score_cal}, we describe how we systematically identify such data using influence functions. 

\subsection{What Counts as High-Quality Robot Data?}
\label{subsec:criteria}
Defining high-quality demonstration data is a challenging problem in data-driven robot learning. 
Existing approaches have adopted varying criteria:
%Traditional approaches vary widely. 
some researchers prioritize optimal behaviors (\textit{e.g.,} shortest path to target objects)~\cite{mandlekar2022matters}, while others emphasize diversity~\cite{ khazatsky2024droid, belkhale2023data}, robustness~\cite{laskey2017dart}.
However, these predefined notions of quality often fail to capture the complex relationship between training data characteristics and final policy performance.
%These definitions may inadequately capture the complex relationship between training data and policy performance.

We propose a more grounded definition: robot data quality should be measured by its \emph{direct performance contribution to the learned policy}. 
Specifically, we quantify a demonstration's quality through its contribution to reducing loss on a small set of validation set consisting of desirable behavior.\footnote{This small validation set can consist of held-out teleoperation data or actual policy rollouts from trained models.}
To calculate this contribution efficiently, we employ influence functions~\cite{koh2017understanding} (detailed in Section~\ref{sec:pre_if}), 
which estimate how validation loss would change if individual training samples were removed, without the computational cost of retraining. Intuitively, understanding how the removal of a training sample affects validation loss allows us to precisely quantify the value of that sample.
% but is this right to argue? since validation loss is calculated on a much smaller portion of the dataset
% defend by rollout definition
% defend with CUPID
We note that from a machine learning perspective, performance on unseen validation samples serves as a natural proxy for a policy's generalization capabilities.
% Also, validation loss can be calculated from successful policy rollouts, and this corresponds to an online reinforcement learning objective using REINFORCE algorithm~\cite{williams1992simple} with return value of 1, which measures the policy performance.
% \textcolor{red}{Also, if successful policy rollouts are used as a validation set, it corresponds to online reinforcement learning objective using REINFORCE algorithm~\cite{williams1992simple} where return value is 1.}
This performance-based definition inherently captures the complex relationship between training data characteristics and policy effectiveness, addressing the limitations of predefined quality criteria.

\subsection{Curating High-Quality Data using Influence Functions}
\label{subsec:score_cal}
Now, we present \Algname (\textbf{\ALGname}), a data curation method based on influence functions.
Formally, given a training dataset $\mathcal{D}_{\tt tr}$ and a trained policy model $\pi_{\theta_{\tt tr}}$, we curate high-quality data using a small validation dataset $\mathcal{D}_{\tt val}$ through the following steps:
    \begin{itemize}
    % \item \textit{Step 1 (Computing influence function)}: 
    \item \textit{Step 1 (Contribution estimation)}: For each state-action pair $(s, a) \in \mathcal{D}_{\tt tr}$, we quantify its contribution to reducing loss on the validation dataset $\mathcal{D}_{\tt val}$ based on influence functions. 
    \item \textit{Step 2 (Trajectory-wise curation)}: We aggregate the influence scores of state-action pairs within each trajectory and select the top $N$ trajectories based on the aggregated scores.
    \end{itemize}
We describe the details of each step in the following paragraphs.

\vspace{4pt}
\textbf{Step 1: Contribution estimation}
For each state-action pair $(s,a) \in \mathcal{D}_{\text{tr}}$, we define
\begin{equation}
g(s,a) := \nabla_\theta \log\pi_{\theta_{\tt tr}}(a|s) / \|\nabla_\theta \log\pi_{\theta_{\tt tr}}(a|s)||,
\end{equation}
and  compute the following score to measure its contribution to reducing loss on the validation dataset $\mathcal{D}_{\tt val}$: 
\begin{equation}
\label{eq:trns_score}
\texttt{\ALGname-score}(s,a) 
:= \max_{(s',a')\in \mathcal{D}_{\tt val}} g(s',a')^{\top} g(s,a).
\end{equation}
% \begin{equation}
%     \label{eq:trns_score}
%      \texttt{\ALGname-score} (s,a) := \max_{(s',a')\in \mathcal{D}_{\tt val}} \nabla'_\theta \log\pi_{\theta_{\tt tr}}(a'|s')^{\top}  \nabla'_{\theta} \log\pi_{\theta_{\tt tr}}(a|s),
% \end{equation}

where $\pi_{\theta_{\tt tr}}$ denotes a policy model trained via BC and $g(s,a)$ is the normalized gradient on log-likelihood. 
% where $\pi_{\theta_{\tt tr}}$ denotes a policy model trained via BC and $\nabla'_\theta  := \frac{\nabla_\theta }{\left\| \nabla_\theta  \right\|_2}$ is the normalized gradient. 
This score, based on the negative of influence functions in \cref{eq:influence_bc_norm}, measures gradient similarity between validation data and training state-action pairs.
% High similarity indicates greater contribution to reducing validation loss, identifying helpful training data.
This similarity indicates the contribution of training state-action pairs to reducing validation loss, where a high score implies helpful training data.

Unlike the original formulation of influence functions that averages gradient products over all validation samples (\textit{i.e.,} $\sum_{(s',a')\in \mathcal{D}_{\tt val}} \nabla_\theta \log\pi_{\theta_{\tt tr}}(a'|s')^{\top}  \nabla'_{\theta} \log\pi_{\theta_{\tt tr}}(a|s)$), we instead take the maximum gradient product across validation samples, which we call \textit{maximum influence scoring}.
% \textcolor{red}{This is because not all validation transitions are helpful to evaluate the quality of state-action pairs of interest, since they could correspond to different behaviors (e.g., pick-and-place behavior might not help identify useful behavior for screwing).
% Therefore, we focus on the most relevant pair by taking the maximum, which helps reduce noise.}
Because each state-action pair in the validation trajectory represents different behaviors, we focus on the most relevant pair by taking the maximum, which helps reduce noise. 
% \textcolor{red}{This is also in contrast to DataMIL~\cite{dass2025datamil} and CUPID~\cite{agia2025cupid}, which calculate influence scores by averaging over validation transitions for robot data curation.}
Our experiments confirm that the maximum influence scoring enhances the reliability of influence estimation, resulting in better performance (see Section~\ref{subsec:ablation} for supporting results).

However, computing and storing gradients in \cref{eq:trns_score} for each sample poses substantial computational challenges for modern robot foundation models with billions of parameters~\cite{black2410pi0, bjorck2025gr00t, kim24openvla}.
To address this, we implement two complementary efficiency strategies: 
First, we selectively compute gradients for only a subset of network layers, specifically excluding parameter-dense components such as vision encoders.
Second, we employ the one-permutation one-random-projection (OPORP) technique~\cite{li2023oporp} to compress gradient vectors while preserving their dot product relationships. 
This compression approach, inspired by recent advances in compute-efficient influence calculation~\cite{kwon2024datainf, min2025understanding}, significantly reduces storage requirements without compromising the accuracy of our scoring mechanism. 

\vspace{4pt}

% \paragraph{Step 2: Trajectory-wise curation}
\textbf{Step 2: Trajectory-wise curation}
After computing the proposed metric $\texttt{\ALGname-score}(s,a)$ in \cref{eq:trns_score}, we aggregate these scores within each trajectory $\tau$ by taking the mean: $\frac{1}{|\tau|} \sum_{(s, a) \in \tau} \texttt{\ALGname-score}(s,a).$
We then select the top $N$ trajectories based on these aggregated scores. 
We adopt trajectory-wise curation specifically to mitigate issues that arise from naively selecting individual state-action pairs with high scores. 
In our preliminary experiments, we observed that 
state-action-wise curation often results in redundant state-action pairs. 
For instance, specific behaviors like grasping moments were disproportionately selected, while other behaviors (such as reaching motions or diverse, multi-modal successful strategies) were filtered out. 
This leads to poor state coverage, which is undesirable for robust policy learning.
By selecting entire trajectories instead, we ensure that the curated dataset maintains diverse state distributions and captures complete behavior sequences. 
Our empirical results in Section~\ref{subsec:ablation} confirm that trajectory-wise curation significantly outperforms state-action-wise
curation across multiple benchmarks.

\section{Experiments}
\label{sec:result}
We design our experiments to answer the following questions:

    \begin{itemize} % TODO : left Margin check
    \item (1) Can \ALGname data curation improve policy success rates? (Section~\ref{subsec:success_rate})
    \item (2) How robust is QoQ to in-the-wild robot data under diverse domains? (Section~\ref{subsec:droid_curation})
    \item (3) Can \ALGname leverage policy rollout as validation set? (Section~\ref{subsec:policy_rollout})
    \item (3) How does each \ALGname component impact performance? (Section~\ref{subsec:ablation})
    \end{itemize}

%%%%%%%%%%%%%%%%%%%%%%%%%%%%%%%%%%%%%%%%%%%%%
\begin{figure*}[t]
    \centering
    \includegraphics[width=0.9\textwidth]{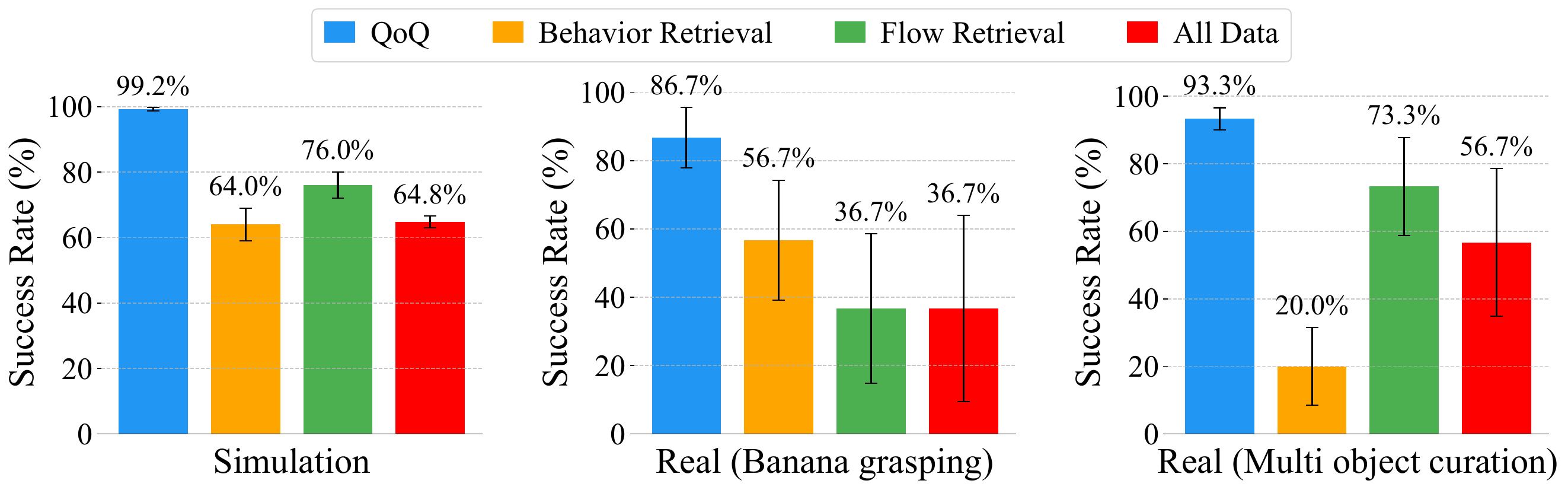}
    \caption{
        \textbf{Success rate for simulation and real robot experiments.} QoQ outperforms all baselines in simulation and real robot experiments by detecting helpful trajectories. We report the mean and standard deviation across 5 runs (simulation) and 3 runs (real robot).
    }
    \label{fig:bc_success_rate}
\end{figure*}
%%%%%%%%%%%%%%%%%%%%%%%%%%%%%%%%%%%

\subsection{Setups}
% \textcolor{red}{We experiment across simulation and real robot setups to test the effectiveness of \ALGname to curate high-quality data.}
\label{subsec:setups}

% \paragraph{Simulation}
\textbf{Environments}
We evaluate \ALGname in both simulation and real-robot setups spanning multiple environments and tasks.
For the simulation experiments, we use the Robomimic benchmark~\cite{mandlekar2022matters}, where a Franka Research 3 robot arm is tasked with placing a Coke can into the correct bin.
% Similar to \cite{du2023behavior}, we use the ``Can-paired" dataset~\cite{mandlekar2022matters} as our training dataset, consisting of 100 successful and 100 failed trajectories, where successful trajectories are characterized by placing the Coke can into the correct bin. 
% \textcolor{red}{By using 10 successful trajectories as a validation set, we aim to curate a dataset that leads to successful task completion by filtering failures.}
% We use 10 successful trajectories as the validation dataset.
% \footnote{The trajectories in the validation dataset involve grasping and moving the can, but do not include the reaching behavior, as both successful and failed trajectories share the same reaching behavior. For the baseline methods, however, we retain the reaching behavior in the validation set, as these methods retrieve individual state-action pairs and perform poorly when the reaching behavior is excluded.} 
\begin{figure}[t]
    \centering
    \includegraphics[width=0.95\columnwidth]{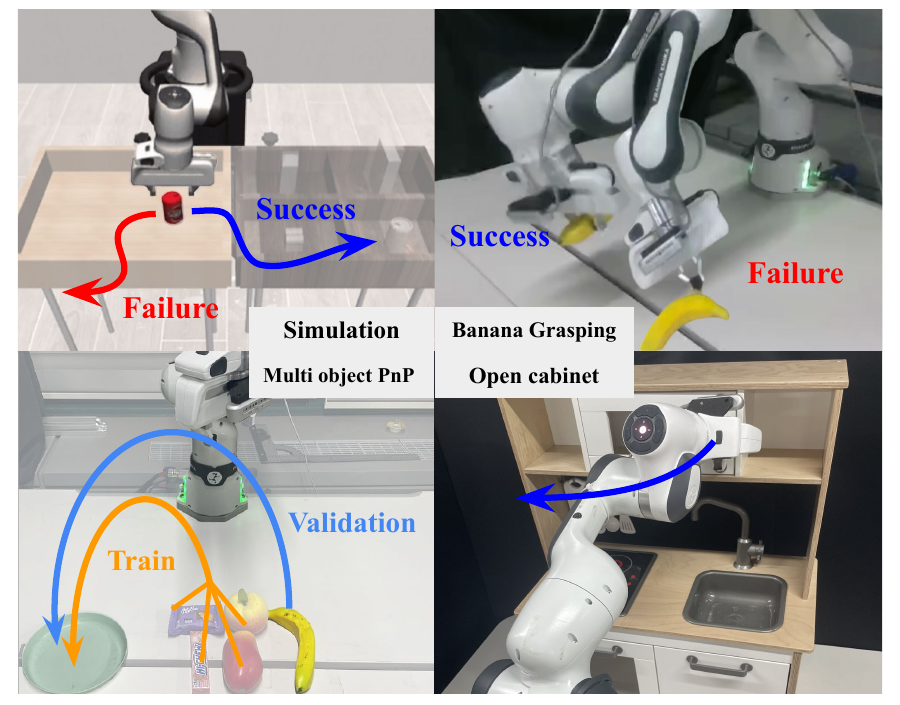}
    \caption{\textbf{Visualization of experiment environments} across simulation and real robot setups, including a single task of grasping a banana and multi-object pick-and-place, and open cabinet task.}
    \vspace{-15pt}
\label{fig:environment_visualization_main}
\end{figure}
% \textbf{Real robot}
% \textcolor{red}{We also evaluate \ALGname on curating a train dataset of multiple real robot tasks.}
For the real-robot experiments, we use a Franka Research 3 robot arm operated via teleoperation to collect training datasets for three tasks: banana grasping, multi-object pick-and-place, and cabinet opening, as shown in Figure~\ref{fig:environment_visualization_main}.
% In real robot experiment, we use a Franka Research 3 robot arm, operated via teleoperation, to collect train datasets for three tasks: a banana grasping experiment, a multi-object pick-and-place experiment, and a cabinet open experiment, as shown in Figure~\ref{fig:environment_visualization_main}. 
% \textcolor{red}{
We describe each task setup in detail sections below.

\textbf{Baselines}
For comparison, we mainly consider the following baselines: 
(1) {\bf All data}: utilizing original demonstrations without any curation.
(2) {\bf Behavior Retrieval}~\cite{du2023behavior}: 
a curation method that retrieves state-action pairs from the training dataset based on their similarity to state-action pairs in the validation dataset. Specifically, it maps state-action pairs from both datasets into a latent space using a Variational Autoencoder (VAE)~\cite{kingma2014auto}, and computes similarity based on distances in the latent space.
(3) {\bf Flow Retrieval}~\cite{linflowretrieval}: 
a curation method similar to Behavior Retrieval that maps optical flow derived from image sequences into a latent space using a VAE, and computes similarity based on distances. Unlike Behavior Retrieval, Flow Retrieval relies solely on optical flow information and does not utilize explicit state-action pairs.
% \vspace{2pt}

\textbf{Metrics}
For evaluation, we measure \textbf{curation accuracy}, defined as the proportion of successful trajectories in the curated dataset.
We also report the \textbf{success rate} of policies trained on the curated datasets.
% \textcolor{red}{We compare the \textbf{success rates} of policies trained on the curated datasets to evalaute performance. We additionally introduce \textbf{curation accuracy}, which is defined as the proportion of successful trajectories in the curated dataset.}
% \vspace{2pt}

\textbf{Experiment details}
For simulated experiments, we use the Transformer policy~\cite{vaswani2017attention} with two self-attention layers with a pretrained ResNet encoder~\cite{he2016deep}.
We train for 1k epochs using a batch size of 100. For real robot experiments, we use GR00T N1~\cite{bjorck2025gr00t}, a high-performing vision-language-action model, and train for 20k steps using LoRA~\cite{hu2022lora} with $\alpha=64$ and rank $r=128$. 
GR00T N1 model is trained via flow matching loss~\cite{lipman2022flow} and we use the gradient of flow matching loss in place of log-likelihood when computing \ALGname scores. This approach has previously been adopted for data attribution in flow matching model classes, as it can be seen as a variational lower bound of the likelihood~\cite{georgiev2023journey, mcallister2025flow}. 
Also, we use gradients from the Transformer blocks in the simulation experiment and from the action-head layers of GR00T N1 in the real robot experiment. In Section~\ref{subsec:ablation}, we examine the impact that varying the curation budget and gradient calculation layers has on the curation accuracy and policy success rates.
% we use flow matching loss to for both training and validation loss when computing \ALGname scores,as it is a regression surrogate for log-likelihood.
% (See Appendix~\ref{subsec:inf_for_diffusion} for details).
Finally, we construct the curated dataset by selecting the top $N$ trajectories with the highest \texttt{\ALGname-score} from the training dataset.
The curation budget $N$ is set to match the number of successful trajectories in the training dataset. If the number of successful trajectories is unknown, we set the curation budget to half the size of the training dataset. For all experiments, validation trajectories are randomly selected.
% For the multi object pick-and-place experiment, as the training dataset does not contain the banana pick-and-place trajectories, while all trajectories are successful regarding the target object, we set the curation budget to be half the size of train dataset.
% For the multi-object pick-and-place experiment, since the entire training dataset consists of successful trajectories and there is no clear criterion for defining a curation budget, we simply set it to half the size of the training dataset.
% The relationship between varying curation budgets and their impact on both curation accuracy and policy success rates is examined in Section~\ref{subsec:ablation}.
% % Appendix~\ref{app:curation_budget}.
% For calculating \texttt{\ALGname-score}, we use gradients from the Transformer blocks in the Simulation experiment and gradients from the action-head layers of GR00T N1 in the real robot experiment. We ablate the choice of gradient calculation layers in \mbox{Section~\ref{subsec:ablation}}.
% Appendix~\ref{app:layer} provides a detailed analysis of how the choice of network layers for gradient computation impacts overall performance.

\subsection{Can \ALGname Data Curation Improve Policy Success Rates?}
\label{subsec:success_rate}
\textbf{Single task curation}
We first evaluate \ALGname and baseline curation methods by curation accuracy in the single task of pick and place of a Coke can in simulation, and the banana grasping experiment in a real robot.

In simulation, similar to Behavior Retrieval~\cite{du2023behavior}, we use the ``Can-paired" dataset~\cite{mandlekar2022matters}  as our train dataset, consisting of 100 successful and 100 failed trajectories, where successful trajectories are characterized by placing the Coke can into the correct bin. 
% The same dataset was also used in Behavior Retrieval baseline~\cite{du2023behavior}.
% We use 10 successful trajectories as the validation dataset.
% \textcolor{red}{randomly chosen}
By using 10 successful trajectories as a validation set, we aim to curate a dataset that leads to successful task completion by filtering failures.\footnote{The trajectories in the validation dataset involve grasping and moving the can, but do not include the reaching behavior, as both successful and failed trajectories share the same reaching behavior. For the baseline methods, however, we retain the reaching behavior in the validation set, as these methods retrieve individual state-action pairs and perform poorly when the reaching behavior is excluded.} 

In the banana grasping experiment, the train dataset consists of 100 real robot trajectories of grasping a banana, with 60 successful and 40 failure trajectories, where failures are characterized by missed grasping points (\textit{e.g.,} attempting to grasp at unsuitable positions or with improper gripper orientations). The validation set consists of 10 successful trajectories.
% \textcolor{red}{held-out successful trajectories which are randomly chosen.}
% In simulation and real robot experiment of grasping banana, we calculate curation accuracy, defined as the percentage of state-action pairs from successful trajectories within the curated dataset.
By taking a curation budget that matches the actual number of successful trajectories in the train data, \ALGname achieves substantially high curation accuracy across both domains, outperforming the best baseline by 31.6\% in simulation and 16.3\% in real robot experiment in the banana grasping experiment as shown in Table~\ref{tab:domain_accuracy}.

\begin{table}[t]
\small
% \captionsetup{font=small}
\centering
\begin{tabular}{l@{\hskip 6pt}c@{\hskip 6pt}c}
\toprule
\textbf{Method} & \textbf{Simulation} & \textbf{Real (Banana grasping)} \\
\midrule
\textbf{\ALGname (Ours)} & \textbf{99.4\stdv{0.3}} & \textbf{83.6\stdv{0.8}} \\
Behavior Retrieval & 67.8\stdv{0.7} & 67.3\stdv{0.7} \\
Flow Retrieval & 56.9\stdv{0.2} & 57.7\stdv{0.1} \\
All Data & 55.4 & 58.7 \\
\bottomrule
\end{tabular}
\caption{\textbf{Curation accuracy (\%).} Percentage of state-action pairs 
from successful trajectories among all pairs in the curated dataset.}
\label{tab:domain_accuracy}
\vspace{-20pt}
\end{table}

We also assess whether these curated datasets translate to improved policy performance. 
We fine-tune policies using datasets curated by \ALGname and baseline methods, then evaluate their success rates during deployment in both simulation and real robot environments. 
As shown in Figure~\ref{fig:bc_success_rate}, policies trained with \ALGname-curated data achieve a 99.2\% success rate in simulation, significantly outperforming Flow Retrieval, the best baseline, which reaches 76.0\%. 
In real robot experiments, \ALGname-trained policies achieve 86.7\% success compared to 56.7\% of the best baseline, Behavior Retrieval.

\vspace{1pt}

% \textbf{Multi task curation} 
\textbf{Multi object curation} 
% \textcolor{blue}{For the multi-object pick-and-place experiment, the training dataset includes 80 trajectories, with 20 each for a peach, mango, snack, and gum, while excluding banana. The validation set consists of 10 successful banana pick-and-place trajectories, and is used to identify the most helpful trajectories for this held-out object.}
In this experiment, we aim to identify helpful trajectories for the pick-and-place of a banana, from a training dataset composed of the pick-and-place of different objects. Specifically, the training dataset includes 80 trajectories, with 20 each for a peach, mango, snack, and gum.
The validation set consists of 10 successful banana pick-and-place trajectories, and is used to identify the most helpful trajectories for this held-out object.
% the training dataset includes pick-and-place demonstrations for multiple objects such as a peach, mango, snack, and gum. 
Unlike the single-task curation experiment described above, this train dataset does not contain explicitly failed trajectories. Rather, we aim to find trajectories that contain helpful information for the pick-and-place of a banana from various tasks.
By constructing the curated set that takes the top 50\% trajectories with the highest \ALGname scores, the downstream policy success rate greatly improves over policy trained in all data, reaching 93.3\% as shown in Figure~\ref{fig:bc_success_rate}.
Meanwhile, Behavior Retrieval completely fails to capture relevant information to help pick-and-place a banana, resulting in the lowest success rate of 20\%\footnote{Note that it is reasonable to achieve a lower success rate than using all data due to the use of fewer trajectories from curation.}. We suspect state-action pair representation from VAEs in Behavior Retrieval is distracted by multiple different objects. Flow Retrieval, meanwhile, performs relatively better by focusing on robot motion captured by optical flow, but still lags behind \ALGname-curated policy. In Appendix~\ref{app:seed_corr}, we also show that \ALGname's data selection is most consistent across different seeds compared to baselines, which clearly shows that \ALGname well captures relevant information for data curation.

%%%%%%%%%%%%%%%%%%%%%%%%%%%%%%%%%%%%%%%%%%%%%%%%%%%%%%%%%%%%%%%%%%%%%%%%%%%%%%%%
% \begin{figure*}[t!]
%     \centering
%     \begin{minipage}{0.48\textwidth}
%         \centering
%         \includegraphics[width=0.8\linewidth]{figures/figure_droid_curation_accuracy.pdf}
%         \caption{\small \textbf{Droid dataset curation accuracy.} Compared to baselines, \ALGname maintains high curation accuracy in DROID dataset, which consists of different domains and object locations.}
%         \label{fig:droid_curation_accuracy}
%     \end{minipage}
%     \hfill
%     \begin{minipage}{0.48\textwidth}
%         \centering
%         \includegraphics[width=\linewidth]{figures/ablation_real_robot_new.pdf}
%         \caption{\small
%         \textbf{Ablation study of QoQ components in our real robot setup.} Removing maximum influence scoring and trajectory-wise curation leads to decreased curation accuracy and policy success rate. Results show the mean and standard deviation across 3 runs.}
%         \label{fig:qoq_ablation}
%     \end{minipage}
% \end{figure*}
%%%%%%%%%%%%%%%%%%%%%%%%%%%%%%%%%%%%%%%%%%%%%%%%%%%%%%%%%%%%%%%%%%%%%%%%%%%%%%%%

\subsection{How Robust Is QoQ to In-the-wild Robot Data under Diverse Domains?}
\label{subsec:droid_curation}
\begin{figure}[t]
    \centering
    \includegraphics[width=0.9\linewidth]{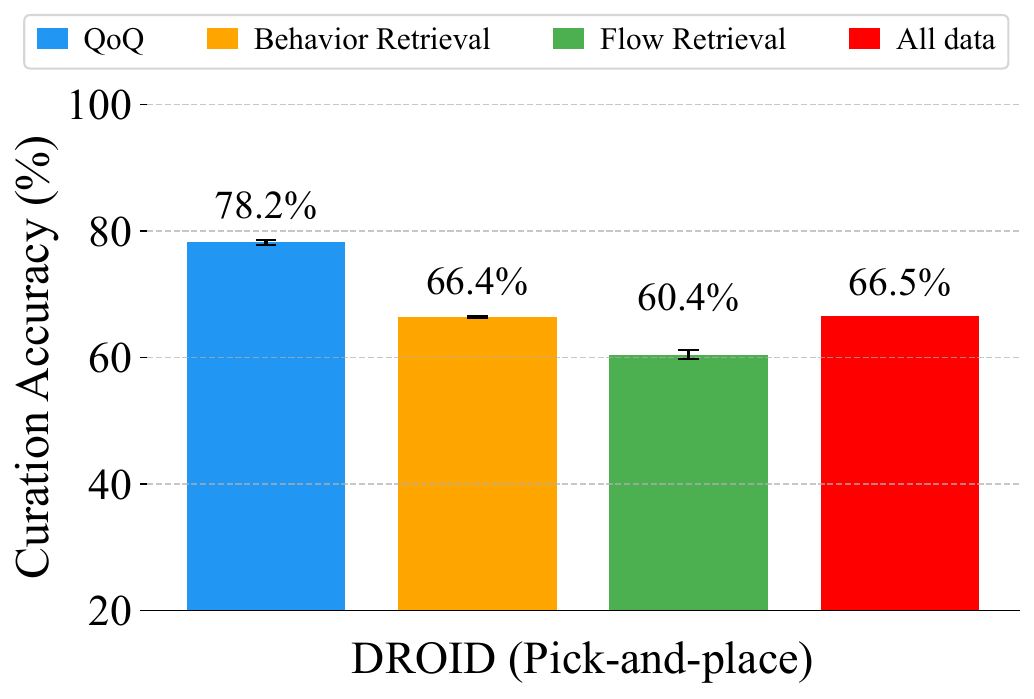}
    \caption{\textbf{Droid dataset curation accuracy (\%).} Compared to baselines, \ALGname maintains high curation accuracy in DROID dataset, which consists of different domains and object locations.}
    \label{fig:droid_curation_accuracy}
    \vspace{-15pt}
\end{figure}

We evaluate the ability of \ALGname to curate high-quality in-the-wild robot data. Specifically, we construct a training dataset using DROID~\cite{khazatsky2024droid}. We sample 200 trajectories of pick-and-place of ``pen/pencil" tasks, which involve 133 successful trajectories and 67 failed trajectories. For the validation set, we use 20 successful trajectories in the same tasks. 
Both successful trajectories and failed trajectories are very challenging to distinguish, as they differ in domains and behavior by varying environments, object location, and camera viewpoints.
For this experiment, we train the GR00T N1 model for 50k steps without LoRA to account for the dataset’s heterogeneity.
% We train GR00T N1 model for 50k steps without LoRA for this experiment to accommodate such heterogeneity in dataset.
To compare different algorithms in this setup, we curate the actual number of successful trajectories and compare the curation accuracy.
% We additionally compare a new baseline DemInf~\cite{hejna2025robot}: a curation method that filters demonstrations negatively impacting mutual information between state and action % (see Appendix~\ref{subsec:deminf_desc} for explanation for DemInf).
Figure~\ref{fig:droid_curation_accuracy} displays the accuracy of the curation in \ALGname, Behavior Retrieval, Flow  Retrieval. 
Our results demonstrate that \ALGname shows the highest curation accuracy in the presence of multiple domains, whereas Behavior Retrieval, Flow Retrieval suffer from training VAE encoders from heterogeneous visual input and diverse behaviors. 
% DemInf also shows poor performance close to random sampling, not being scaled to the complex datasets.

\subsection{Can \ALGname Use Policy Rollout as Validation Set?}
\label{subsec:policy_rollout}
Instead of assuming a separate validation set in addition to the train data, we experiment with using trajectories acquired by policy rollout as a validation set when curating train data.
% In this section, we experiment with curating train dataset using 
% This setup removes the need for additional creation of validation data for data curation. 
For example, a policy trained on an uncurated dataset can be used to collect a validation set for data curation.
% While we assume that the validation set of \ALGname involves ``desirable behavior", 
However, policy rollouts often involve failures due to poor initial performance, while we define validation to contain only desirable behavior. To make use of the failed trajectories as a validation set, we use the negative value of \ALGname score after calculating the score using the failed validation trajectories. 
% justfication
This intuitively makes sense because for a specific trajectory in the train dataset, if \ALGname score is high when using failed trajectories as a validation set, it implies that the trajectory encourages such failed behavior, so it should be discouraged.

%\textcolor{red}{
We calculate \ALGname score the same as before for successful policy rollout and subtract the \ALGname score calculated from failed trajectories. To balance between two scores, we weight each score by the size of the validation set used, as it can account for the uncertainty of scores coming from different sizes of validation sets.
%}
% Since we are calculating two positive and negative QoQ scores using either successful or failed trajectories as a validation set, the weight between both scores remains a design choice. Empirically, we find that using the size of the validation set used to calculate the score as a weight is effective, as it can account for the confidence of score according to the validation set size. 
% Still, we can leverage failed trajectories to curate train dataset, as it can provide information about the bad behavior that leads to task failure.
% This is often useful as it provides information on bad behavior that leads to failure.
% However, such an autonomous rollout often involves task failure, which is often useful as it provides information on bad behavior that leads to the failure. Therefore, when evaluating train dataset using a failed dataset as a validation set, we use the negative of \ALGname score to curate the dataset.
To verify this setup, we collect a training dataset on the ``Open cabinet" task, including 100 successful and 50 failure trajectories. As shown in Figure~\ref{fig:environment_visualization_main}, this task involves grasping and pulling the cabinet door handle to open it. We also vary the cabinet location to 5 different positions to maximize task difficulty. 
% During evaluation, we assign a score of 0.25 for reaching the cabinet handle, 0.5 for grasping it, 0.75 for pulling it, and 1 (task success) for opening the cabinet by more than 30 degrees. 
After training our policy on the training dataset, we conducted 20 rollouts from the policy and acquired 5 successful and 15 failure trajectories.

\begin{wrapfigure}[16]{r}{0.22\textwidth}
    \centering
    \vspace{-10pt}
    \includegraphics[width=\linewidth]{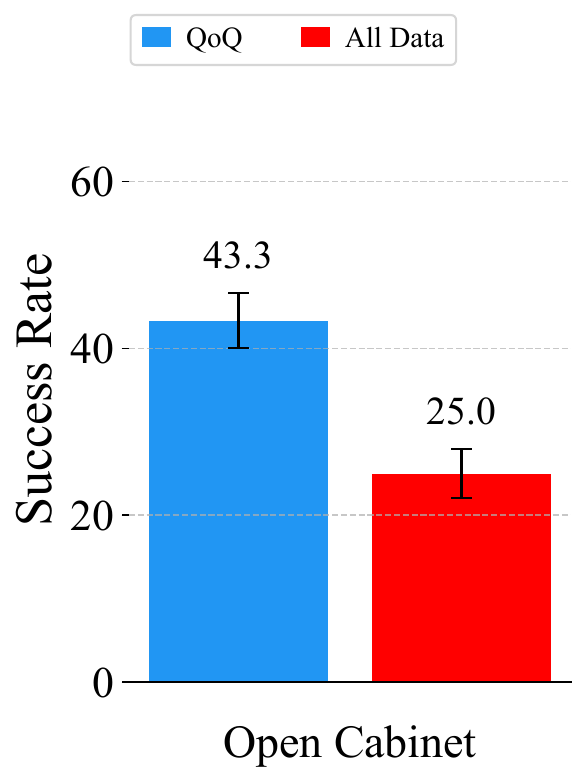}
    \caption{\textbf{Open cabinet policy success rate.}}
    \label{fig:open_cabinet}
\end{wrapfigure}

\vspace{5pt}

%\textbf{Calculating QoQ score}
% \textcolor{red}{
From the training dataset, we take 100 trajectories where such weighted QoQ scores are the highest to construct a curated dataset.
% }
% In addition to our original baselines, we introduce a new baseline CUPID~\cite{agia2025cupid}, which also leverages policy rollouts as the validation set.
% Our curated dataset achieves the curation accuracy of
% \input{tables/curation_accuracy}
% As shown here, \ALGname successfully detects successful trajectories in train dataset resulting \% higher percentage of success in the curated set. 
% \textcolor{red}{
Then, we fine-tune policy using this curated set and evaluate policy success rate in real robot deployment. As shown in Figure~\ref{fig:open_cabinet}, the policy trained using the curated dataset achieves a higher policy success rate compared to the policy trained on all data. This shows that \ALGname can also improve the performance of the pretrained policy without assuming initial availability for the validation set.
% }
% This broadens the application of \ALGname. 
% \ALGname improves the performance 
% \ALGname achieves the highest policy score of 74.2\%, showing that \ALGname can effectively use policy rollouts as a validation set. It outperforms CUPID, which also uses influence functions to curate the training dataset. We believe that the difference in performance stems from multiple techniques from \ALGname, especially maximum influence scoring, which captures the most relevant state-action pair from the validation set.
% Figure~\ref{fig:open_cabinet} shows the score of the policy trained using the curated dataset from each method in addition to successful rollout. \ALGname achieves the highest policy score of 74.2\%, showing that \ALGname can effectively use policy rollouts as a validation set. It outperforms CUPID, which also uses influence functions to curate the training dataset. We believe that the difference in performance stems from multiple techniques from \ALGname, especially maximum influence scoring, which captures the most relevant state-action pair from the validation set.
% \textbf{Policy rollout data curation}
% We also verify whether QoQ remains effective when using rollout data generated by a trained policy, rather than teleoperation data.

\subsection{How Does Each Component in \ALGname Impact Performance?}
\label{subsec:ablation}

In this section, we present the ablation study of \ALGname, analyzing the contribution of each component to the overall performance.
All experiments are conducted on the banana grasping experiment task using real robots.

\vspace{5pt}
% \paragraph{Maximum influence scoring}

\begin{figure}[t]
    \centering
    \includegraphics[width=0.95\linewidth]{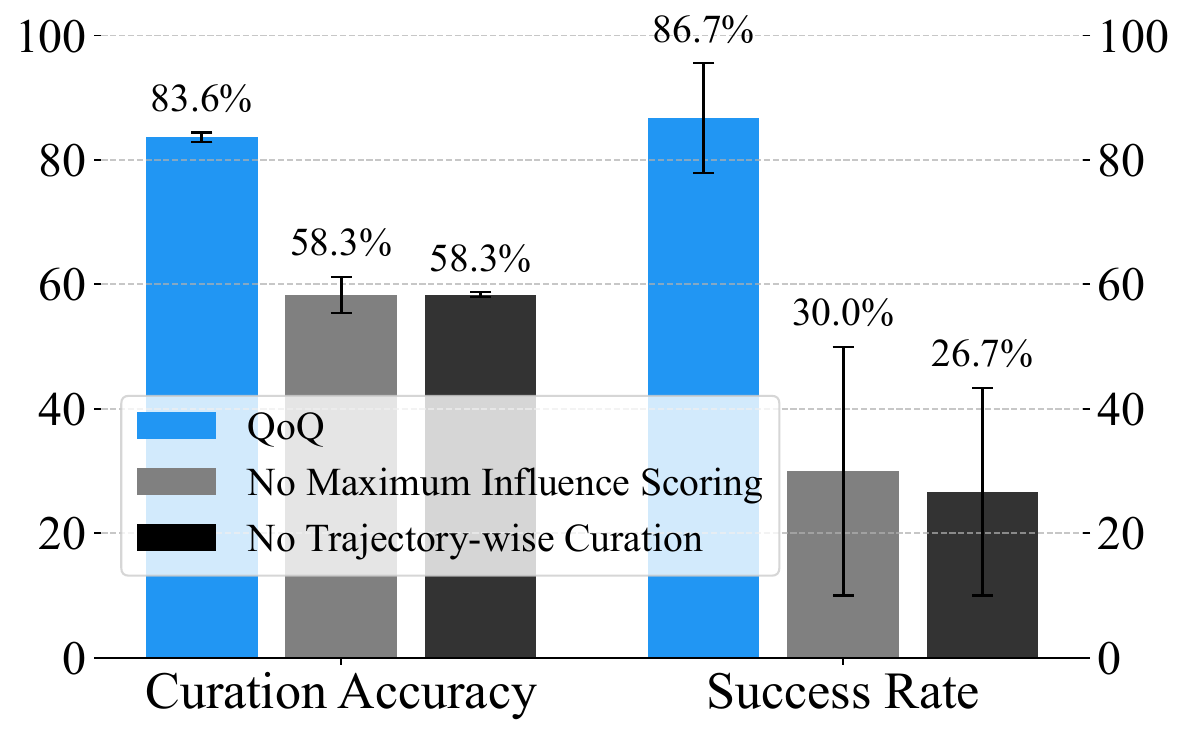}
    \caption{
    % \small 
    \textbf{Ablation study of QoQ components} in banana grasping experiment. Removing maximum influence scoring and trajectory-wise curation reduces curation accuracy and policy success rate. Error bars show mean$\pm$stderr over 3 runs.
    }
    \label{fig:qoq_ablation}
    \vspace{-20pt}
\end{figure}

\textbf{Maximum influence scoring.}
As discussed in Step 1 of Section~\ref{subsec:score_cal}, \ALGname computes scores by taking the maximum gradient product across validation samples, instead of averaging gradients over all validation samples. 
To demonstrate the effectiveness of this approach, we compare both curation accuracy and the success rate of policies trained on datasets curated using our maximum influence scoring against those curated by an alternative approach that averages the gradient products.
As shown in Figure~\ref{fig:qoq_ablation}, our method achieves higher accuracy and success rates. 
This improvement arises from its ability to ignore irrelevant validation samples when scoring each training sample.

% \paragraph{Trajectory-wise curation}
\textbf{Trajectory-wise curation.}
Similarly, we compare the success rates between trajectory-wise curation (as discussed in Step 2 of Section~\ref{subsec:score_cal}) and state-action-wise curation.
Figure~\ref{fig:qoq_ablation} shows that trajectory-wise curation results in higher success rates. 
This gain is due to its ability to mitigate distributional bias in the curated dataset, which could otherwise degrade the performance of the BC policy during training.

\vspace{2pt}

\textbf{Gradient computation layer}
We compare curation accuracy when computing the \ALGname score from different network layers of the GROOT N1 model~\cite{bjorck2025gr00t}.
Table~\ref{tab:grad_layer} shows that applying \ALGname to only a subset of network modules yields results that are consistent with those obtained from the full parameters. 
This suggests that effective influence estimation does not require computing gradients across all layers.
% , but instead remains robust when restricted to informative components such as the action head. 
As a result, \ALGname is particularly well-suited for scaling to modern VLAs~\cite{kim24openvla, bjorck2025gr00t, black2410pi0}, which involve billions of parameters.
\begin{table}[H]
\small
% \captionsetup{font=small}

\centering
\captionsetup{skip=10pt} 
\begin{tabular}{lccc}
\toprule
\textbf{Layer Part} & \textbf{Backbone} & \textbf{Action Head} & \textbf{All} \\
\midrule
\textbf{Curation Accuracy} & 82.7\stdv{0.6} & \bf{83.6}\stdv{1.3} & 82.1\stdv{1.2} \\
\bottomrule
\end{tabular}
\caption{\textbf{Curation accuracy across different influence computation layer parts} in banana grasping experiment. Backbone denotes the VLM and vision encoder part used in GROOT N1 model.}
\label{tab:grad_layer}
\vspace{-10pt}
\end{table}
% In our original experiment, we selectively computed gradients for only a subset of network layers to reduce computation overhead. We used transformer block in the simulation experiment and action head block in real robot experiment for gradient computation. Specifically, transformer block denotes two self-attention layers of transformer policy, and action head block consists of state encoder, action encoder, diffusion transformer block, and action decoder in GROOT N1 model (see Section~\ref{sec:exp_detail} for details).
\vspace{-2pt}
\textbf{Curation budget}
We present experimental results on how varying the curation budget affects both curation accuracy and downstream policy performance.
When the number of curated trajectories is set below or near the proportion of successful trajectories in the training dataset, we find that curated policies consistently achieve substantially higher task success rates compared to policies trained without curation.
Across a wide range of curation budgets, our method also outperforms the strongest baseline, Behavior Retrieval, demonstrating its effectiveness (Table~\ref{tab:curation_budget}).
However, the curation budget still has a non-negligible impact on final policy performance, as smaller budgets reduce data coverage.
Thus, in practice, we recommend exploring multiple curated set sizes to maximize policy performance.

\begin{table}[H]
% \captionsetup{font=small}
\small
% \captionsetup{font=small}
\captionsetup{skip=6pt}
\setlength{\tabcolsep}{3pt}
\centering
\begin{tabular}{lcccc@{\hspace{6pt}\vrule\hspace{6pt}}cc}
\toprule
\makecell{\textbf{Curation}\\\textbf{Budget}} & \textbf{10} & \textbf{20} & \textbf{40} & \textbf{60}
& \makecell{\textbf{All}\\\textbf{data}}
& \makecell{\textbf{Best}\\\textbf{baseline}} \\
\midrule
\makecell{\textbf{Success}\\\textbf{Rate (\%)}} 
& \makecell{36.7\\ \stdv{17.6}}
& \makecell{63.3\\ \stdv{17.6}}
& \makecell{60.0\\ \stdv{10.0}}
& \makecell{\bfseries 86.7\\ \stdv{8.9}}
& \makecell{36.7\\ \stdv{27.3}}
& \makecell{56.7\\ \stdv{17.6}} \\
\bottomrule
\end{tabular}
\caption{\textbf{Success rates across curation budgets.} \ALGname consistently achieves higher success rates across a wide range of budgets compared to baselines. Best baseline refers to the Behavior Retrieval method.}
\label{tab:curation_budget}
\vspace{-10pt}
\end{table}

%%%%%%%%%%%%%%%%%%%%%%%%%%%%%%%%%%%%%%%%%%%%%%%%%%%%%%%%%%%%%%%%%%%%%%%%%%%%%%%%
% \begin{table}[h]
% \caption{An Example of a Table}
% \label{table_example}
% \begin{center}
% \begin{tabular}{|c||c|}
% \hline
% One & Two\\
% \hline
% Three & Four\\
% \hline
% \end{tabular}
% \end{center}
% \end{table}

%    \begin{figure}[thpb]
%       \centering
%       \framebox{\parbox{3in}{We suggest that you use a text box to insert a graphic (which is ideally a 300 dpi TIFF or EPS file, with all fonts embedded) because, in an document, this method is somewhat more stable than directly inserting a picture.
% }}
%       %\includegraphics[scale=1.0]{figurefile}
%       \caption{Inductance of oscillation winding on amorphous
%        magnetic core versus DC bias magnetic field}
%       \label{figurelabel}
%    \end{figure}
%%%%%%%%%%%%%%%%%%%%%%%%%%%%%%%%%%%%%%%%%%%%%%%%%%%%%%%%%%%%%%%%%%%%%%%%%%%%%%%%

\section{CONCLUSIONS}

In this work, we propose \ALGname, a method that curates robotic datasets based on direct performance contribution to the learned policy.
We define a quality scoring mechanism for state-action pairs derived from influence functions.
To enhance curation effectiveness and policy performance, we introduce two key components:
(1) maximum influence scoring and (2) trajectory-wise curation. 
Our experiments demonstrate that \ALGname effectively identifies high-quality robotic trajectories, resulting in significantly improved policy success rates compared to baseline curation algorithms across both simulation and real-world robot experiments. Also, \ALGname reliably detects success and failed trajectories and curates helpful trajectories from the in-the-wild DROID datasets.
We believe that our scoring method offers a promising approach for data-driven robot learning, enabling more efficient use of demonstration data to achieve high-performing robotic policies.

\textbf{Limitations and Future Work}
Our approach still has several limitations that suggest promising directions for future work. While trajectory-level curation ensures broad coverage, it cannot selectively use high-quality segments within a trajectory, motivating finer-grained sub-trajectory curation. Influence function computation remains costly and approximate, despite layer restrictions and methods like TracIn ~\cite{pruthi2020estimating} or OPORP~\cite{li2023oporp}, calling for more accurate yet efficient estimators. Moreover, our setup assumes shared embodiments between training and validation, whereas extending to cross-embodiment scenarios (e.g., Open X-Embodiment~\cite{o2024open}) would broaden applicability. Although we focus on behavioral cloning, the method can generalize to other policy objectives, such as offline RL.

%%%%%%%%%%%%%%%%%%%%%%%%%%%%%%%%%%%%%%%%%%%%%%%%%%%%%%%%%%%%%%%%%%%%%%%%%%%%%%%%

%%%%%%%%%%%%%%%%%%%%%%%%%%%%%%%%%%%%%%%%%%%%%%%%%%%%%%%%%%%%%%%%%%%%%%%%%%%%%%%%

%%%%%%%%%%%%%%%%%%%%%%%%%%%%%%%%%%%%%%%%%%%%%%%%%%%%%%%%%%%%%%%%%%%%%%%%%%%%%%%%
\section*{APPENDIX}

\subsection{Data Selection Consistency Experiment}
\label{app:seed_corr}
% \textcolor{red}{
In this section, we compare the data selection consistency between \ALGname and baselines across from multi-object curation experiment across three different seeds. % Specifically, we use the train dataset used in multi-object curation experiment that consists of the pick and place of peach, mango, snack, and gum in Section~\ref{subsec:success_rate}. 
% We then train the policy using three different random seeds and compare the resulting trajectory rankings. 
Our hypothesis is that more consistent rankings across seeds indicate that the algorithm provides more reliable trajectory scores. Specifically, we use Kendall's W coefficient~\cite{abdi2007kendall} to compare the correlation of ranking.
\begin{table}[H]
\centering
\captionsetup{skip=10pt} 
\small
\renewcommand{\arraystretch}{1.2}
\begin{tabular}{lccc}
\toprule
\textbf{Metric} & \textbf{QoQ} & \textbf{Behavior Retrieval} & \textbf{Flow Retrieval} \\
\midrule
Kendall's W & 0.7713 & 0.3287 & 0.7026 \\
\bottomrule
\end{tabular}
\caption{Kendall’s W values across different methods. Higher indicates more consistent trajectory rankings.}
\label{tab:rank_correlation}
\vspace{-15pt}
\end{table}
Table~\ref{tab:rank_correlation} shows that \ALGname shows the most consistent ranking across seeds, followed by Flow Retrieval and Behavior Retrieval algorithms. This also matches the success rates of downstream policies shown in Section~\ref{subsec:success_rate}, demonstrating that \ALGname well captures relevant information for data curation.
%}

\section*{ACKNOWLEDGMENT}
%\textcolor{red}{
The authors would like to thank Changyeon Kim, Juyong Lee and Dongjun Lee for providing helpful comments for improving the work. This work was supported by Institute for Information \& communications Technology Planning \& Evaluation(IITP) grant funded by the Korea government(MSIT) (RS-2019-II190075, Artificial Intelligence Graduate School Program(KAIST)),  the Korea government(MSIT) (No. RS-202400509279, Global AI Frontier Lab), and Institute of Information \& Communications Technology Planning \& Evaluation(IITP) grant (RS-2025-02304967, AI Star Fellowship(KAIST)) funded by the Korea government(MSIT).
%}
% \section*{ACKNOWLEDGMENT}
% The preferred spelling of the word ÒacknowledgmentÓ in America is without an ÒeÓ after the ÒgÓ. Avoid the stilted expression, ÒOne of us (R. B. G.) thanks . . .Ó  Instead, try ÒR. B. G. thanksÓ. Put sponsor acknowledgments in the unnumbered footnote on the first page.

%%%%%%%%%%%%%%%%%%%%%%%%%%%%%%%%%%%%%%%%%%%%%%%%%%%%%%%%%%%%%%%%%%%%%%%%%%%%%%%%

\bibliographystyle{IEEEtran}
\bibliography{IEEEabrv}

@inproceedings{koh2017understanding,
  title={Understanding black-box predictions via influence functions},
  author={Koh, Pang Wei and Liang, Percy},
  booktitle={International Conference on Machine Learning},
  year={2017},
}

@article{du2023behavior,
  title={Behavior retrieval: Few-shot imitation learning by querying unlabeled datasets},
  author={Du, Maximilian and Nair, Suraj and Sadigh, Dorsa and Finn, Chelsea},
  journal={arXiv preprint arXiv:2304.08742},
  year={2023}
}

@inproceedings{linflowretrieval,
  title={FlowRetrieval: Flow-Guided Data Retrieval for Few-Shot Imitation Learning},
  author={Lin, Li-Heng and Cui, Yuchen and Xie, Amber and Hua, Tianyu and Sadigh, Dorsa},
  booktitle={Conference on Robot Learning},
  year={2024}
}

@inproceedings{mandlekar2022matters,
  title={What Matters in Learning from Offline Human Demonstrations for Robot Manipulation},
  author={Mandlekar, Ajay and Xu, Danfei and Wong, Josiah and Nasiriany, Soroush and Wang, Chen and Kulkarni, Rohun and Fei-Fei, Li and Savarese, Silvio and Zhu, Yuke and Mart{\'\i}n-Mart{\'\i}n, Roberto},
  booktitle={Conference on Robot Learning},
  year={2022},
}

@inproceedings{pruthi2020estimating,
  title={Estimating training data influence by tracing gradient descent},
  author={Pruthi, Garima and Liu, Frederick and Kale, Satyen and Sundararajan, Mukund},
  booktitle={Advances in Neural Information Processing Systems},
  year={2020}
}

@article{min2025understanding,
  title={Understanding Impact of Human Feedback via Influence Functions},
  author={Min, Taywon and Lee, Haeone and Ryu, Hanho and Kwon, Yongchan and Lee, Kimin},
  journal={arXiv preprint arXiv:2501.05790},
  year={2025}
}

@inproceedings{pomerleau1988alvinn,
  title={Alvinn: An autonomous land vehicle in a neural network},
  author={Pomerleau, Dean A},
  booktitle={Advances in Neural Information Processing Systems},
  year={1988}
}

@inproceedings{o2024open,
  title={Open x-embodiment: Robotic learning datasets and rt-x models: Open x-embodiment collaboration},
  author={O’Neill, Abby and Rehman, Abdul and Maddukuri, Abhiram and Gupta, Abhishek and Padalkar, Abhishek and Lee, Abraham and Pooley, Acorn and Gupta, Agrim and Mandlekar, Ajay and Jain, Ajinkya and others},
  booktitle={IEEE International Conference on Robotics and Automation},
  year={2024},
}

@inproceedings{walke2023bridgedata,
  title={Bridgedata v2: A dataset for robot learning at scale},
  author={Walke, Homer Rich and Black, Kevin and Zhao, Tony Z and Vuong, Quan and Zheng, Chongyi and Hansen-Estruch, Philippe and He, Andre Wang and Myers, Vivek and Kim, Moo Jin and Du, Max and others},
  booktitle={Conference on Robot Learning},
  year={2023},
}

@article{kingma2014auto,
  title={Auto-Encoding Variational Bayes},
  author={Kingma, Diederik P and Welling, Max},
  journal={arXiv preprint arXiv:1312.6114},
  year={2014}
}

@article{hejna2025robot,
  title={Robot Data Curation with Mutual Information Estimators},
  author={Hejna, Joey and Mirchandani, Suvir and Balakrishna, Ashwin and Xie, Annie and Wahid, Ayzaan and Tompson, Jonathan and Sanketi, Pannag and Shah, Dhruv and Devin, Coline and Sadigh, Dorsa},
  journal={arXiv preprint arXiv:2502.08623},
  year={2025}
}

@article{khazatsky2024droid,
  title={Droid: A large-scale in-the-wild robot manipulation dataset},
  author={Khazatsky, Alexander and Pertsch, Karl and Nair, Suraj and Balakrishna, Ashwin and Dasari, Sudeep and Karamcheti, Siddharth and Nasiriany, Soroush and Srirama, Mohan Kumar and Chen, Lawrence Yunliang and Ellis, Kirsty and others},
  journal={arXiv preprint arXiv:2403.12945},
  year={2024}
}

@inproceedings{
  nasiriany2022learning,
  title={Learning and Retrieval from Prior Data for Skill-based Imitation Learning},
  author={Soroush Nasiriany and Tian Gao and Ajay Mandlekar and Yuke Zhu},
  booktitle={Conference on Robot Learning},
  year={2022},
}

@inproceedings{belkhale2023data,
  title={Data quality in imitation learning},
  author={Belkhale, Suneel and Cui, Yuchen and Sadigh, Dorsa},
  booktitle={Advances in Neural Information Processing Systems},
  year={2023}
}

@inproceedings{ghorbani2019data,
  title={Data shapley: Equitable valuation of data for machine learning},
  author={Ghorbani, Amirata and Zou, James},
  booktitle={International Conference on Machine Learning},
  year={2019},
}

@article{kwon2021beta,
  title={Beta shapley: a unified and noise-reduced data valuation framework for machine learning},
  author={Kwon, Yongchan and Zou, James},
  journal={arXiv preprint arXiv:2110.14049},
  year={2021}
}

@article{jia2019efficient,
  title={Efficient task-specific data valuation for nearest neighbor algorithms},
  author={Jia, Ruoxi and Dao, David and Wang, Boxin and Hubis, Frances Ann and Gurel, Nezihe Merve and Li, Bo and Zhang, Ce and Spanos, Costas J and Song, Dawn},
  journal={arXiv preprint arXiv:1908.08619},
  year={2019}
}

@inproceedings{
  kwon2024datainf,
  title={DataInf: Efficiently Estimating Data Influence in Lo{RA}-tuned {LLM}s and Diffusion Models},
  author={Yongchan Kwon and Eric Wu and Kevin Wu and James Zou},
  booktitle={International Conference on Learning Representations},
  year={2024},
}

@article{black2410pi0,
  title={$\pi$0: A vision-language-action flow Model for general robot control},
  author={Black, Kevin and Brown, Noah and Driess, Danny and Esmail, Adnan and Equi, Michael and Finn, Chelsea and Fusai, Niccolo and Groom, Lachy and Hausman, Karol and Ichter, Brian and others},
  journal={arXiv preprint arXiv:2410.24164},
  year={2024}
}

@article{bjorck2025gr00t,
  title={GR00T N1: An Open Foundation Model for Generalist Humanoid Robots},
  author={Bjorck, Johan and Casta{\~n}eda, Fernando and Cherniadev, Nikita and Da, Xingye and Ding, Runyu and Fan, Linxi and Fang, Yu and Fox, Dieter and Hu, Fengyuan and Huang, Spencer and others},
  journal={arXiv preprint arXiv:2503.14734},
  year={2025}
}

@article{kim24openvla,
    title={OpenVLA: An Open-Source Vision-Language-Action Model},
    author={{Moo Jin} Kim and Karl Pertsch and Siddharth Karamcheti and Ted Xiao and Ashwin Balakrishna and Suraj Nair and Rafael Rafailov and Ethan Foster and Grace Lam and Pannag Sanketi and Quan Vuong and Thomas Kollar and Benjamin Burchfiel and Russ Tedrake and Dorsa Sadigh and Sergey Levine and Percy Liang and Chelsea Finn},
    journal = {arXiv preprint arXiv:2406.09246},
    year={2024},
}

@article{team2024octo,
  title={Octo: An open-source generalist robot policy},
  author={Team, Octo Model and Ghosh, Dibya and Walke, Homer and Pertsch, Karl and Black, Kevin and Mees, Oier and Dasari, Sudeep and Hejna, Joey and Kreiman, Tobias and Xu, Charles and others},
  journal={arXiv preprint arXiv:2405.12213},
  year={2024}
}

@inproceedings{li2023oporp,
  title={OPORP: One permutation+ one random projection},
  author={Li, Ping and Li, Xiaoyun},
  booktitle={ACM SIGKDD Conference on Knowledge Discovery and Data Mining},
  year={2023}
}

@inproceedings{he2016deep,
  title={Deep residual learning for image recognition},
  author={He, Kaiming and Zhang, Xiangyu and Ren, Shaoqing and Sun, Jian},
  booktitle={IEEE Conference on Computer Vision and Pattern Recognition},
  year={2016}
}

@inproceedings{
  hu2022lora,
  title={Lo{RA}: Low-Rank Adaptation of Large Language Models},
  author={Edward J Hu and yelong shen and Phillip Wallis and Zeyuan Allen-Zhu and Yuanzhi Li and Shean Wang and Lu Wang and Weizhu Chen},
  booktitle={International Conference on Learning Representations},
  year={2022},
}

@inproceedings{memmel2025strap,
  title={{STRAP}: Robot Sub-Trajectory Retrieval for Augmented Policy Learning},
  author={Marius Memmel and Jacob Berg and Bingqing Chen and Abhishek Gupta and Jonathan Francis},
  booktitle={International Conference on Learning Representations},
  year={2025},
}

@article{chen2025curating,
  title={Curating Demonstrations using Online Experience},
  author={Chen, Annie S and Lessing, Alec M and Liu, Yuejiang and Finn, Chelsea},
  journal={arXiv preprint arXiv:2503.03707},
  year={2025}
}

@inproceedings{laskey2017dart,
  title={Dart: Noise injection for robust imitation learning},
  author={Laskey, Michael and Lee, Jonathan and Fox, Roy and Dragan, Anca and Goldberg, Ken},
  booktitle={Conference on Robot Learning},
  year={2017},
}

@inproceedings{vaswani2017attention,
  title={Attention is all you need},
  author={Vaswani, Ashish and Shazeer, Noam and Parmar, Niki and Uszkoreit, Jakob and Jones, Llion and Gomez, Aidan N and Kaiser, {\L}ukasz and Polosukhin, Illia},
  booktitle={Advances in Neural Information Processing Systems},
  year={2017}
}

@article{dass2025datamil,
  title={DataMIL: Selecting Data for Robot Imitation Learning with Datamodels},
  author={Dass, Shivin and Khaddaj, Alaa and Engstrom, Logan and Madry, Aleksander and Ilyas, Andrew and Mart{\'\i}n-Mart{\'\i}n, Roberto},
  journal={arXiv preprint arXiv:2505.09603},
  year={2025}
}

@article{abdi2007kendall,
  title={The Kendall rank correlation coefficient},
  author={Abdi, Herv{\'e}},
  journal={Encyclopedia of measurement and statistics},
  volume={2},
  pages={508--510},
  year={2007},
  publisher={Sage Thousand Oaks, CA}
}

@article{agia2025cupid,
  title={CUPID: Curating Data your Robot Loves with Influence Functions},
  author={Agia, Christopher and Sinha, Rohan and Yang, Jingyun and Antonova, Rika and Pavone, Marco and Nishimura, Haruki and Itkina, Masha and Bohg, Jeannette},
  journal={arXiv preprint arXiv:2506.19121},
  year={2025}
}

@article{georgiev2023journey,
  title={The journey, not the destination: How data guides diffusion models},
  author={Georgiev, Kristian and Vendrow, Joshua and Salman, Hadi and Park, Sung Min and Madry, Aleksander},
  journal={arXiv preprint arXiv:2312.06205},
  year={2023}
}

@article{lipman2022flow,
  title={Flow matching for generative modeling},
  author={Lipman, Yaron and Chen, Ricky TQ and Ben-Hamu, Heli and Nickel, Maximilian and Le, Matt},
  journal={arXiv preprint arXiv:2210.02747},
  year={2022}
}

@article{mcallister2025flow,
  title={Flow Matching Policy Gradients},
  author={McAllister, David and Ge, Songwei and Yi, Brent and Kim, Chung Min and Weber, Ethan and Choi, Hongsuk and Feng, Haiwen and Kanazawa, Angjoo},
  journal={arXiv preprint arXiv:2507.21053},
  year={2025}
}

\end{document}